\documentclass[american,a4paper, 10pt, conference]{ieeeconf}
\pdfoutput=1
\usepackage[T1]{fontenc}
\usepackage[latin9]{inputenc}
\usepackage{array}
\usepackage{amsmath}
\usepackage{amssymb}
\usepackage{graphicx}
\usepackage{esint}

\makeatletter

\providecommand{\tabularnewline}{\\}

 \usepackage[ruled,vlined]{algorithm2e}

\IEEEoverridecommandlockouts   

\usepackage{xcolor}
\usepackage{bm}

\usepackage{times}

\@ifundefined{showcaptionsetup}{}{%
 \PassOptionsToPackage{caption=false}{subfig}}
\usepackage{subfig}
\makeatother

\begin{document}

\title{Finding Near-optimal Solutions in Multi-robot Path Planning}

\author{Michal \v{C}\'{a}p$^{1}$, Peter Nov\'{a}k$^{2}$, Alexander Kleiner$^{3}$\thanks{$^1$ Agent Technology Center, Dept. of Computer Science, Faculty of Electrical Engineering, CTU in Prague}
\thanks{$^2$ Algorithmics, EEMCS, Delft University of Technology}
\thanks{$^3$ iRobot Inc., Pasadena}}
\maketitle
\begin{abstract}
We deal with the problem of planning collision-free trajectories for
robots operating in a shared space. Given the start and destination
position for each of the robots, the task is to find trajectories
for all robots that reach their destinations with minimum total cost
such that the robots will not collide when following the found trajectories.
Our approach starts from individually optimal trajectory for each
robot, which are then penalized for being in collision with other
robots. The penalty is gradually increased and the individual trajectories
are iteratively replanned to account for the increased penalty until
a collision-free solution is found. Using extensive experimental evaluation,
we find that such a penalty method constructs trajectories with near-optimal
cost on the instances where the optimum is known and otherwise with
4-10\,\% lower cost than the trajectories generated by prioritized
planning and up to 40\,\% cheaper than trajectories generated by
local collision avoidance techniques, such as ORCA.

\end{abstract}

\section{Introduction}

Thanks to recent advances in robotics, teams of autonomous robotic
systems start to pervade not only industrial settings, but also our
private spaces. Apart from numerous deployments in military settings~\cite{selecky_2013_aamas_demo},
autonomously navigating robots are more and more found in warehouses
and manufacturing plants \cite{Wurman_KIVA_IAAI07}, many private
households make use of autonomous cleaning robots, and transport systems
consisting of large numbers of unmanned areal vehicles (UAVs) are
considered for delivering medium sized packages to the consumers. 

One of the important problems in multi-robotics is simultaneous operation
of multiple autonomous vehicles in shared spaces and their mutual
collision avoidance. That is, given a number of vehicles, together
with their starting and destination positions, we are interested in
finding a set of individual trajectories so that the do not collide
with each other, while at the same time the overall costs (e.g., sum
of trajectory lengths) is minimized.

The problem of finding such trajectories has been studied under different
names since at least 1980s. It is known that even the simple variant
of the problem involving rectangular-shaped vehicles in a bounded
2-d space is PSPACE-hard~\cite{hopcroft84}.  While the problem
is relatively straightforward to formulate as a planning problem in
the Cartesian product of the state spaces of the individual robots,
the solutions are difficult to find using standard search techniques
because the joint state-space grows exponentially with the number
of robots. The complexity can be partly mitigated using independence
detection techniques such as ID~\cite{Standley10} and M{*}~\cite{WagnerC11_Mstar}
which detect independent conflict clusters and solve the resulting
lower-dimensional problems separately. However, each such sub-conflict
can be still prohibitively large to solve. 

Prioritized planning~\cite{Erdmann87onmultiple,cap_2012_adpp} is
a heuristic approach based on the idea of sequential planning for
the individual robots in the order of their priorities, where each
robot considers the trajectories of higher-priority robots as moving
obstacles and plans its trajectory to avoid them. While fast, prioritized
planning is incomplete and often fails to find a solution even if
one exists. Moreover, due to its greedy nature, the resulting trajectories
are typically noticeably suboptimal.

Reactive approaches based on the velocity obstacle paradigm such as
DRCA~\cite{lalish2011distributed} or ORCA~\cite{vanBerg2011_ORCA}
are also popular in practice thanks to their computational efficiency.
However, they resolve conflicts only locally and thus they cannot
guarantee that the resulting motion will be deadlock-free.

Efficient guaranteed algorithms such as Push\&Rotate~\cite{deWilde_push_and_rotate_aamas}
and Bibox~\cite{Surynek:2009:NAP:1703435.1703586} exist for a specific
formulation of the problem, where the robots move on a graph, and
each robot occupies exactly one vertex. A collision occurs only if
two robots occupy the same vertex or travel on the same edge. Hence,
these techniques are not suitable for fine-grained planning of robot's
motion, where the collision is defined in terms of the separation
distance between the robots.

The techniques of mathematical optimization have been also studied
in the context trajectory generation. In particular, the penalty-based
approaches~\cite{nocedal_wright_numerical_optimization} have been
tried for single-robot trajectory generation \cite{Schulman_RSS13_Finding_Locally_Optimal_Trajectories_with_SQP}
and for the rendezvous multi-robot planning problem~\cite{BhattacharyaRSS10}. 

We explore the penalty-based approach in the context of multi-robot
path finding for collision avoidance and introduce the \emph{k-step
penalty method} that can be seen as a generalization of prioritized
planning approach. The approach solves the multi-robot path planning
problem by performing a series of single-robot path planning queries
in a dynamic environment in which trajectories that get close to trajectories
of other robots are penalized. That is, starting from trajectories
that disregard collisions with other robots, the conflicts between
robots' trajectories are gradually being penalized with increasing
severity so as to finally, in a limit, the trajectories of individual
robots are forced out of conflict regions as the penalties tend to
infinity. Using extensive experiments, we demonstrate that this heuristic
approach tends to generate near-optimal trajectories that are of significantly
lower cost than the trajectories generated by currently used techniques
for collision avoidance prioritized planning and ORCA~\cite{vanBerg2011_ORCA}.

After a brief problem statement in the following section, Section~\ref{sec:Algorithm}
introduces the kPM algorithm. Subsequently, Section~\ref{sec:Experimental-Evaluation}
provides an extensive experimental evaluation and analysis of the
algorithm's performance and compares it with the relevant state-of-the-art
algorithms in terms of success rate, solution quality and runtime.
Finally, Section~\ref{sec:Conclusion} concludes the paper with a
discussion of its contributions.

\section{\label{sec:Multi-robot-Path-Planning}Multi-robot Path Planning}

Consider a team of circular mobile robots indexed $1,\ldots,n$, each
with a radius $r_{i}>0$ operating in a shared 2-d workspace with
static obstacles. Each robot has a task to move from its start position
$s_{i}$ to some goal position $g_{i}$. A trajectory of robot $\pi_{i}$
is a mapping $\pi_{i}(t):\,\left[0,\infty\right)\rightarrow\mathbb{R}^{2}$
representing the position of the center of the robot at each future
time point. A trajectory of each robot is required to start at the
robot's starting position $s_{i}$ and finish at its goal position
$g_{i}$. Further, each such trajectory $\pi_{i}$ bears a cost, denoted
$c(\pi_{i})$. For simplicity, we will identify the cost of a trajectory
with the time the robot spends outside its destination position. 

We say that two trajectories $\pi_{i},\pi_{j}$ are \emph{conflict-free}
iff the bodies of the two robots never intersect during the execution
of their trajectories. More formally, for robots $i$ and $j$ we
require that for each time-point $t\in\left[0,\infty\right)$, the
distance of the corresponding robots' positions is greater than the
sum of their radii, formally:
\[
\forall t\in\left[0,\infty\right):\:\left|\pi_{i}(t),\pi_{j}(t)\right|>r_{i}+r_{j},
\]
where $\left|\cdot,\cdot\right|$ is the Euclidean distance between
two points. 

Given the assumptions above, the \emph{multi-robot path planning problem}
is to find a set of trajectories $\pi_{1}^{*},\ldots,\pi_{n}^{*}$
corresponding to the individual robots $1,\ldots,n$, such that i)
each pair of trajectories $\pi_{i},\pi_{j}$ with $i\neq j$ is conflict-free,
and ii) the sum of trajectory costs $\sum_{i=1}^{n}c(\pi_{i})$ is
minimal.

To solve the problem, we could consider all robots in the system as
one composite robot with many degrees of freedom and use some path
planning algorithm to find a joint path for all the robots. However,
the size of such a joint configuration space grows exponentially with
the number of robots and thus this approach quickly becomes impractical
if one wants to plan for more than a few robots. 

A pragmatic approach that is often useful even for large multi-robot
teams is prioritized planning. The idea has been first articulated
by Erdman and Lozano-P\'{e}rez in \cite{Erdmann87onmultiple}. In prioritized
planning each robot is assigned a unique priority. The trajectories
for individual robots are then planned sequentially from the highest
priority robot to the lowest priority one. For each robot a trajectory
is planned so that it avoids both the static obstacles in the environment
as well as the higher-priority robots moving along the trajectories
planned in the previous iterations. Works such as \cite{BergO05,Bennewitz02Planning}
investigate heuristics for choosing a good priority sequence for the
robots. Prioritized planning is more efficient than planning in the
joint configuration space, but there are scenarios in which prioritized
planning fails to provide a solution even if all possible priority
sequences are tried (cf.~Figure~\ref{fig:Instance_where_PP_fails}).
Further, the solutions generated by prioritized planning are in most
cases noticeably suboptimal, see Figure~\ref{fig:Instance_where_PP_is_suboptimal}
for an example of such scenario.

\begin{center}
\begin{figure}[t]
\begin{centering}
\includegraphics[scale=0.6]{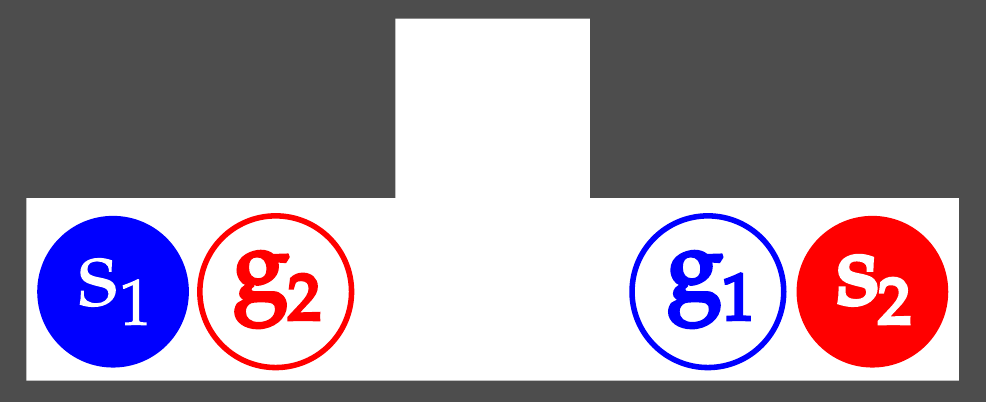}
\par\end{centering}

\caption{\label{fig:Instance_where_PP_fails}Corridor swap scenario: The picture
shows two robots desiring to move from $s_{1}$ to $g_{1}$ ($s_{2}$
to $g_{2}$) in a corridor that is only slightly wider than a body
of a single robot. Both robots move at identical maximum speeds. Irrespective
of which robot starts planning first, its trajectory will be in conflict
with all satisfying trajectories of the second robot. }
\end{figure}

\par\end{center}

\begin{center}
\begin{figure}[t]
\begin{centering}
\subfloat[Optimal solution]{\begin{centering}
\includegraphics[scale=0.6]{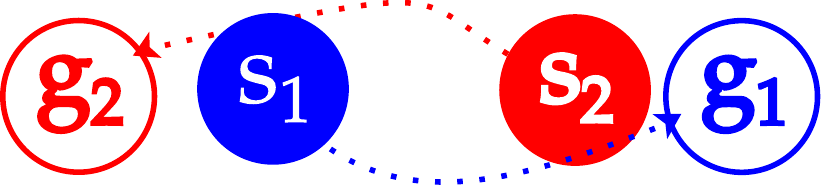}
\par\end{centering}

}~\\
~\\

\par\end{centering}

\begin{centering}
\subfloat[Solution found by prioritized planing]{\begin{centering}
\includegraphics[scale=0.55]{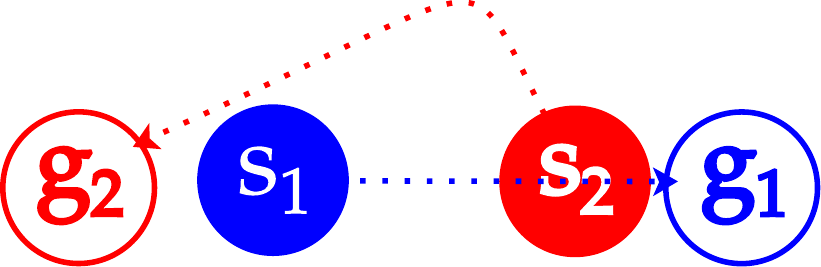}
\par\end{centering}

}
\par\end{centering}

\caption{\label{fig:Instance_where_PP_is_suboptimal}Heads-on scenario: The
picture shows two robots desiring to move from $s_{1}$ to $g_{1}$
($s_{2}$ to $g_{2}$ resp.). The top picture illustrates how an optimal
solution looks like. The bottom picture shows a solution generated
by prioritized planning assuming that the robot $1$ has the higher
priority: robot $1$ will follow a straight line path to its destination,
without considering robot $2$, which will have to bear the full cost
of avoiding the collision.}
\end{figure}

\par\end{center}

\section{\label{sec:Algorithm}Penalty-based Method}

We propose an approach which can be seen as a generalization of prioritized
planning that attempts to mitiagate the two mentioned drawbacks of
the prioritized approach, but in the same time retain its tractability.
We combine the idea of decoupled planning as used in prioritized planning
with a process of iterative increasing of penalty, which is a popular
approach for solving constrained optimization problems~\cite{nocedal_wright_numerical_optimization}. 

In the proposed approach, the requirement on minimal separation between
two trajectories is modeled by a penalty function that assigns a penalty
to each pair of trajectories based on how much do they violate the
separation requirement.  The solution to the multi-robot pathfinding
problem is constructed by gradually increasing the penalty assigned
when a robot passes through a collision region and by letting each
robot replan its trajectory to account for the increased penalty.

Initially, the algorithm ignores interactions between the robots and
finds an optimal trajectory for each robot. Then, the algorithms starts
gradually increasing penalty. After each increase in weight, one of
the robots is selected and a new optimal trajectory that reflects
the increased penalty is found for the robot. The process is repeated
until the penalties are large enough to start dominating the cost
of trajectories, effectively forcing them out of the collision regions. 

\begin{figure}[t]
\subfloat[\label{fig:Bump-function}Three bump functions with parameters $p_{\mathit{max}}=1$,
$d_{ij}^{\mathit{sep}}=1$ and $s=1/3$ (blue), $s=1$ (purple) and
$s=3$ (yellow)]{\begin{centering}
\includegraphics[width=0.45\columnwidth]{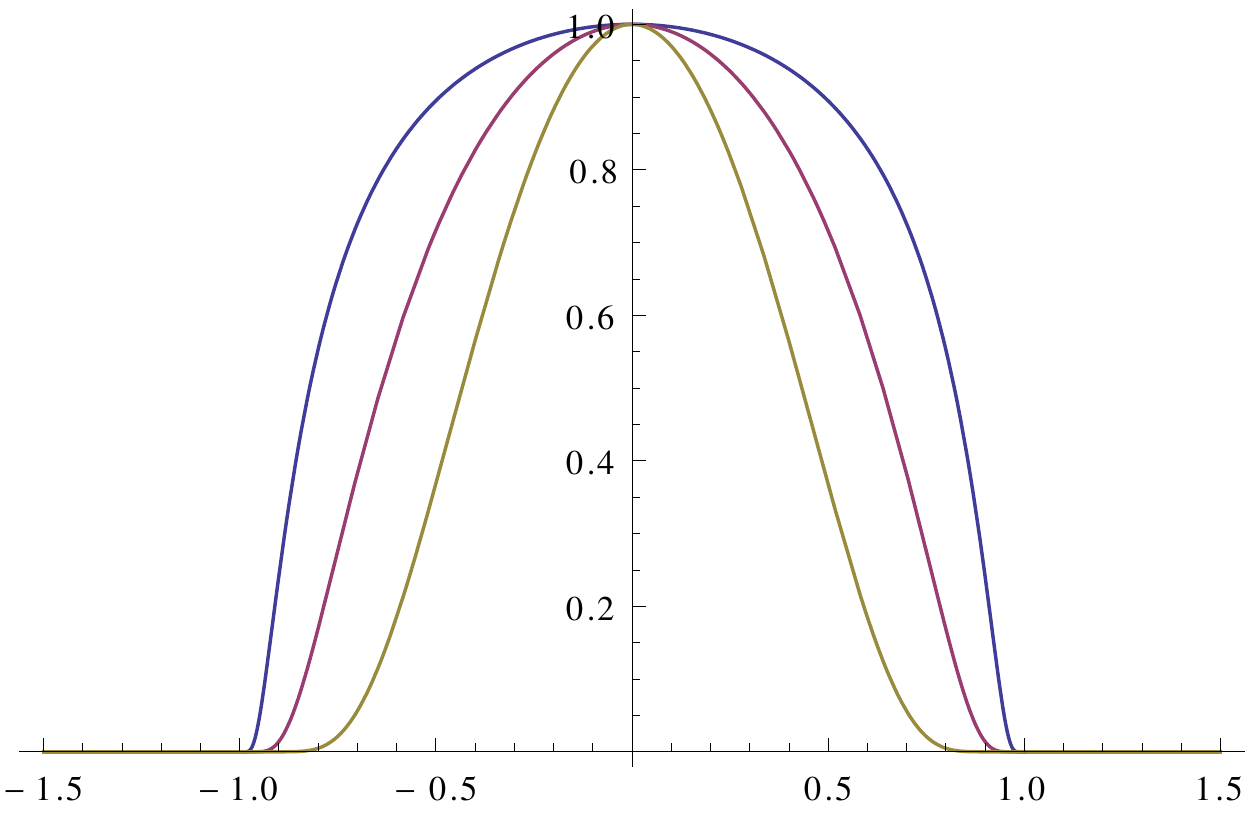}
\par\end{centering}

}~~~~\subfloat[\label{fig:tan-seq}Example 30 elements long weight sequence $w_{1},\ldots,w_{30}$ ]{\begin{centering}
\includegraphics[width=0.45\columnwidth]{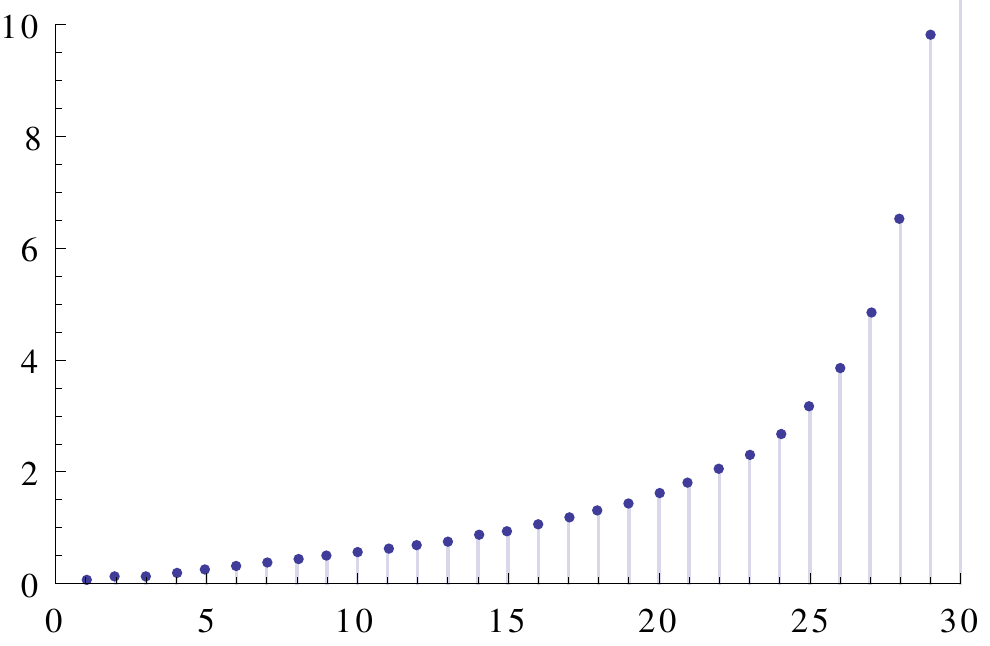}
\par\end{centering}

}

\caption{Penalty function and weight sequence function}
\end{figure}

The minimal separation constraints between a pair of robots $i,j$
is approximated by a penalty function assigning penalty to each part
of the trajectory of robot $i$ that gets closer to the trajectory
of robot $j$ than the required separation distance $d_{ij}^{sep}=r_{i}+r_{j}$.
The penalty function has the form 
\[
\Omega_{ij}(\pi_{i},\pi_{j})=\int_{0}^{\infty}\omega_{ij}(\left|\pi_{i}(t)-\pi_{j}(t)\right|)\, dt\,,
\]
where $\omega_{ij}(d):\,\mathbb{R}_{\geq0}\rightarrow\mathbb{R}_{\geq0}$
is a continuous function assigning a time-point penalty for  interaction
of the trajectories $\pi_{i}$ and $\pi_{j}$ at a time-point $t$.
In the case the trajectories do not violate the separation distance
$d_{sep}$ at the timepoint $t$, the function assigns zero penalty.
Generally, the function needs to satisfy the following conditions:
\[
\forall d\enskip\begin{cases}
\omega_{ij}(d)=0 & \text{ if }d\geq d_{ij}^{sep}\\
\omega_{ij}(d)>0 & \text{ if }d<d_{ij}^{sep}
\end{cases}.
\]
An example of a smooth function that satisfies these conditions is
a bump function 
\[
\omega_{ij}(d)=\begin{cases}
\frac{p_{\mathit{max}}}{e^{-s}}\cdot e^{-\frac{s}{1-(d/d_{ij}^{\mathit{sep}})^{2}})} & \textrm{for }d<d_{ij}^{\mathit{sep}}\\
0 & \mathrm{\textrm{otherwise}}
\end{cases}\:,
\]

where $p_{\mathit{max}}$ is a constant that can be used to adjust
the penalty at $d=0$ and $s$ is a constant that can be used to adjust
the ``steepness'' of the function. The intuition is that the ``more''
the vehicles violate the separation constraint in a given time-point,
that is, the closer they get, the higher penalty should be assigned
to the violation in that time-point. Figure~\ref{fig:Bump-function}
depicts example plots of three bump functions having different values
of the parameter $s$. 

\SetKwProg{alg}{Algorithm}{}{}
\SetKwProg{procedure}{Procedure}{}{}
\SetKwProg{function}{Function}{}{}
\SetKwFunction{replan}{Replan}
\SetKwFunction{pm}{PM}
\LinesNumbered
\begin{algorithm}[t]
\alg{\pm{k}}{

	\For{$i\gets1\ldots n$}{

		$\pi_{i}\gets$\replan{$i,0$}\;\label{alg:kpm:initial}

	}

	\For{$i\gets1\ldots n(k-2)$}{

		 $r\leftarrow i\mathrm{\: mod}\: n(k-2)$\; \label{alg:local-iterative:round_robin-1}

		 $w_{i}\leftarrow\tan(\frac{i}{n(k-2)+1}\cdot\frac{\pi}{2})$\;

		$\pi_{i}\gets$\replan{$i,w_{i}$}\;\label{alg:kpm:iterative}

	}

	\For{$i\gets1\ldots n$}{

		$\pi_{i}\gets$\replan{$i,\infty$}\;\label{alg:kpm:final}

	}

	\uIf{$\forall_{ij}\:\Omega_{ij}(\pi_{i},\pi_{j})=0$}{

		\Return$\left\langle \pi_{1},\ldots,\pi_{n}\right\rangle $\;

	}\Else{

		report failure\;

	}

}

\function{\replan{$r,w$}}{

	return trajectory $\pi$ for robot $r$ that minimizes $c(\pi)+\underset{j\neq r}{w\sum\,}\Omega_{rj}(\pi,\pi_{j})$
\;

}

\caption{\label{alg:kPM}k-step Penalty Method}
\end{algorithm}

Algorithm~\ref{alg:kPM} exposes the \emph{k-step Penalty Method}
(kPM) algorithm that replans the trajectory of each robot exactly
$k$-times. The algorithm starts by finding a cost-optimal trajectory
for each robot using $w=0$, i.e., while ignoring interactions with
other robots. Then, it gradually increases the weight $w$ and thus
the penalties start to be taken into account. After each increase
of the weight coefficient, one of the robots is selected and its trajectory
is replanned to account for the increased penalty. After the iterative
phase finishes, the trajectories of all robots are replanned for the
last time with the weight coefficient set to infinity. If the final
set of trajectories is conflict-free, the algorithm returns the trajectories
as a valid solution, otherwise it reports failure. 

Since each robot replans once in the beginning (line~\ref{alg:kpm:initial})
and once at the end (line~\ref{alg:kpm:final}) of the algorithm,
the robot is left with $k-2$ opportunities for replanning in the
iterative phase of the algorithm. Hence, it performs $n(k-2)$ iterations
in the iterative stage with all the robots taking turns in a round-robin
fashion (line~\ref{alg:kpm:iterative}).

During the iterative phase, the weights are increased in a sequence
$w_{1},\ldots,w_{l}$, with $l=n(k-2)$. To model the gradual hardening
of the separation constraints ultimately to the hard separation constraint,
the sequence needs to converge to infinity. One way of generating
such a sequence is using functions of the following form $w_{i}=\tan(\frac{i}{l+1}\cdot\frac{\pi}{2})$
for $i=1,\ldots,l$. Figure~\ref{fig:tan-seq} shows an example plot
of such a sequence. 

In each iteration of the algorithm, a selected robot can generally
respond in two ways to a weight increase. On the one hand side, it
can accept the increase and simply leave its trajectory unchanged.
On the other hand side, in can find a higher-cost trajectory that
avoids the penalized region. The optimal trajectory for the robot,
however, often lies in between these two extremes and avoids the penalty
region only partially such that the increased cost of the trajectory
with decreased received penalty are optimally traded off. This corresponds
to finding a optimal trajectory subject to a spatio-temporal cost
function. To find such a trajectory we discretize the free space in
form of a grid-like graph and add a discretized time dimension. An
optimal discrete solution is then obtained by running the A{*} algorithm
on such a space-time graph.

\subsection*{}

\begin{figure}
\begin{centering}
\subfloat[The graph used for trajectory planning (in gray).]{\begin{centering}
\includegraphics[width=0.48\columnwidth]{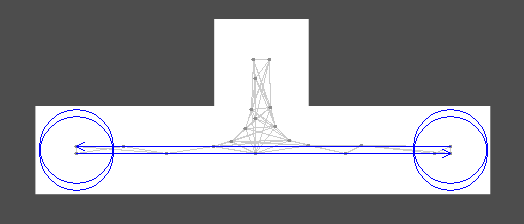}
\par\end{centering}

}\subfloat[Initial trajectories]{\begin{centering}
\includegraphics[width=0.48\columnwidth]{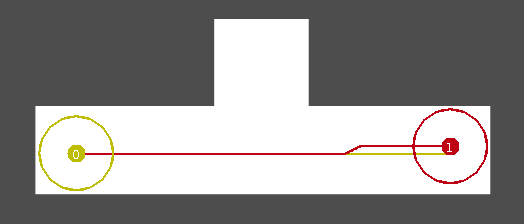}
\par\end{centering}

}
\par\end{centering}

\begin{centering}
\subfloat[Trajectories at iteration 3]{\begin{centering}
\includegraphics[width=0.48\columnwidth]{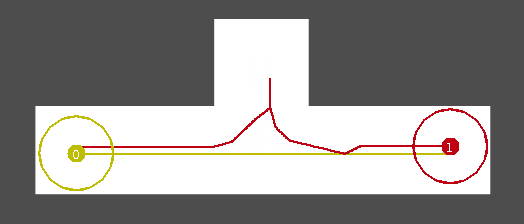}
\par\end{centering}

}\subfloat[Trajectories at iteration 9. Final.]{\begin{centering}
\includegraphics[width=0.48\columnwidth]{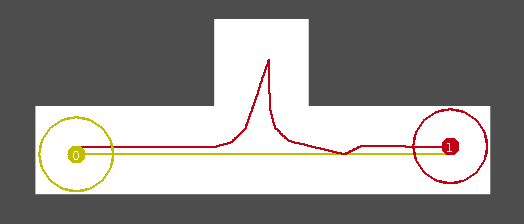}
\par\end{centering}

}
\par\end{centering}

\caption{\label{fig:Swap-in-corridor}Corridor swap scenario: two robot have
to swap their positions in the depicted narrow corridor. The frames
show intermediate solutions generated by PM(k=10). A video showing
the resolution process and simulated execution of the solution can
be watched at http://youtu.be/HartJqN5HXM.}
\end{figure}
\begin{figure}
\begin{centering}
\subfloat[The graph used for trajectory planning (in gray).]{\begin{centering}
\includegraphics[width=0.48\columnwidth]{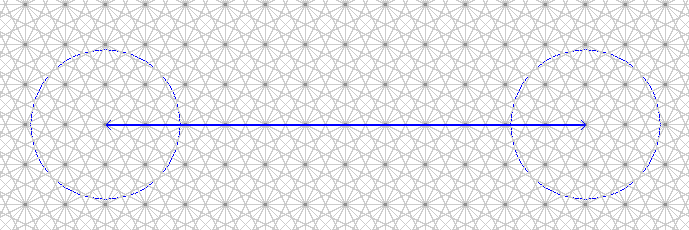}
\par\end{centering}

}\subfloat[Initial trajectories.]{\begin{centering}
\includegraphics[width=0.48\columnwidth]{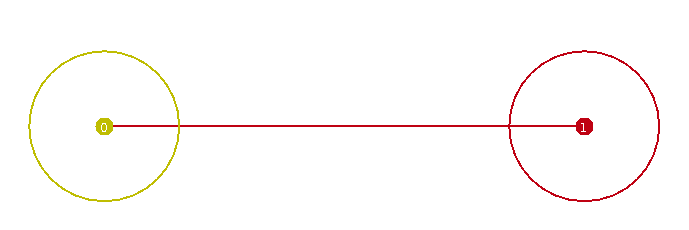}
\par\end{centering}

}
\par\end{centering}

\begin{centering}
\subfloat[Trajectories at iteration 19.]{\begin{centering}
\includegraphics[width=0.48\columnwidth]{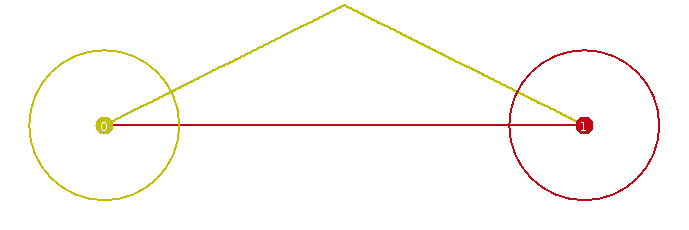}
\par\end{centering}

}\subfloat[Trajectories at iteration 20.]{\begin{centering}
\includegraphics[width=0.48\columnwidth]{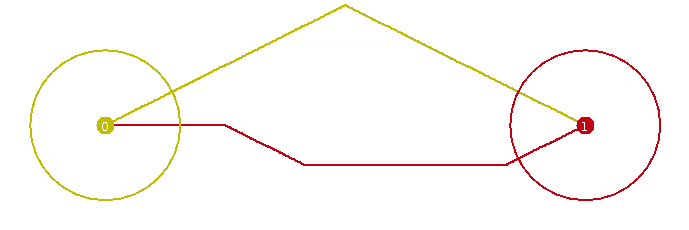}
\par\end{centering}

}
\par\end{centering}

\begin{centering}
\subfloat[Trajectories at iteration 21.]{\begin{centering}
\includegraphics[width=0.48\columnwidth]{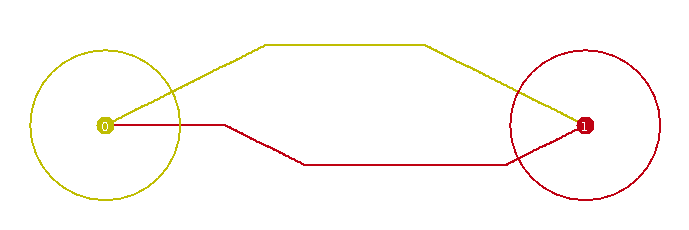}
\par\end{centering}

}\subfloat[Trajectories at iteration 23.]{\begin{centering}
\includegraphics[width=0.48\columnwidth]{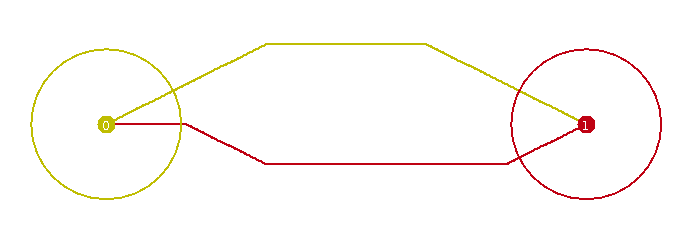}
\par\end{centering}

}
\par\end{centering}

\centering{}\subfloat[Trajectories at iteration 30. Final. (Cost: 21.84)]{\begin{centering}
\includegraphics[width=0.48\columnwidth]{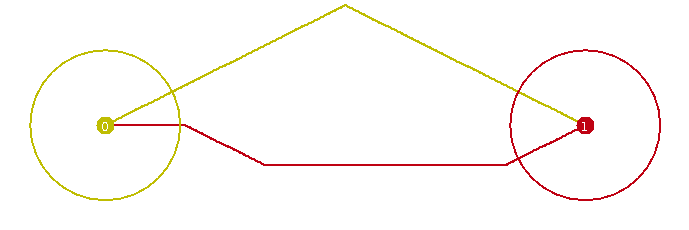}
\par\end{centering}

}\subfloat[Solution found by prioritized planning. (Cost: 22.68)]{\begin{centering}
\includegraphics[width=0.45\columnwidth]{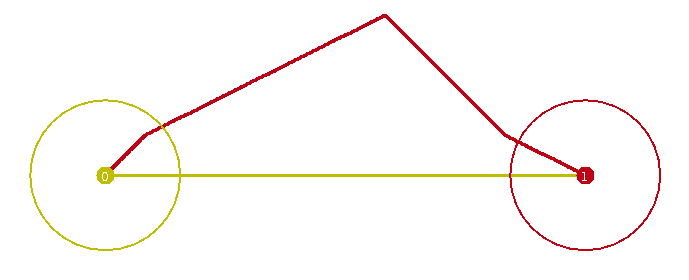}
\par\end{centering}

}\caption{\label{fig:Heads-on-scenario}Heads-on scenario: the two robots need
to swap their position in empty space. The frames show intermediate
solutions generated by PM(k=20). A video showing the resolution process
and simulated execution of the solution can be watched at http://youtu.be/E7LFPIi3zhQ.}
\end{figure}

\subsection*{Illustration of benefits}

The penalty method has two benefits over prioritized planning. First,
the penalty method can solve instances that are out of reach of prioritized
planning. For example, the corridor swap scenario (introduced in Figure~\ref{fig:Instance_where_PP_fails})
can be resolved by the penalty method in $k=10$ iterations. Figure~\ref{fig:Swap-in-corridor}
illustrates how penalty method resolves the corridor swap problem.
Second, for many problem instances, the penalty method can find cheaper
solutions than prioritized planning. For instance, if the penalty
method is applied to the heads-on scenario (introduced in Figure~\ref{fig:Instance_where_PP_is_suboptimal}),
the algorithm constructs a solution in which both robots slightly
divert theirs trajectories and divide the cost of collision avoidance.
This solution is cheaper then the solution returned by prioritized
planning in which one of the robots takes all the cost. Figure~\ref{fig:Heads-on-scenario}
illustrates how penalty method resolves the heads-on problem.

\section{\label{sec:Experimental-Evaluation}Experimental Evaluation}

In this section we compare the performance of our k-step penalty method
(abbreviated as PM or PM(k=$\ldots$)) against prioritized planning
(PP) and a state-of-the-art optimal algorithm called operator decomposition
(OD) on a range of dense multi-robot path planning instances. Specifically,
we focus on small and dense collision situations in which all robots
are involved in a single conflict cluster. These situations usually
represent bottlenecks in solving multi-robot pathfinding problems
since sparser scenarios can be in most cases decomposed into a number
of independent conflict clusters and solved separately. 

Since reactive techniques based on the velocity obstacle paradigm~\cite{fiorini1998_velocity_obstacle}
are often used as a practical approach for collision avoidance between
robots in multi-robot teams, we also compare our method with optimal
reciprocal collision avoidance (ORCA), one of the most popular algorithms
belonging to this family. 

\begin{figure*}
\begin{centering}
\begin{tabular}{>{\centering}m{5cm}>{\centering}m{5cm}>{\centering}m{5cm}}
\includegraphics[width=4cm]{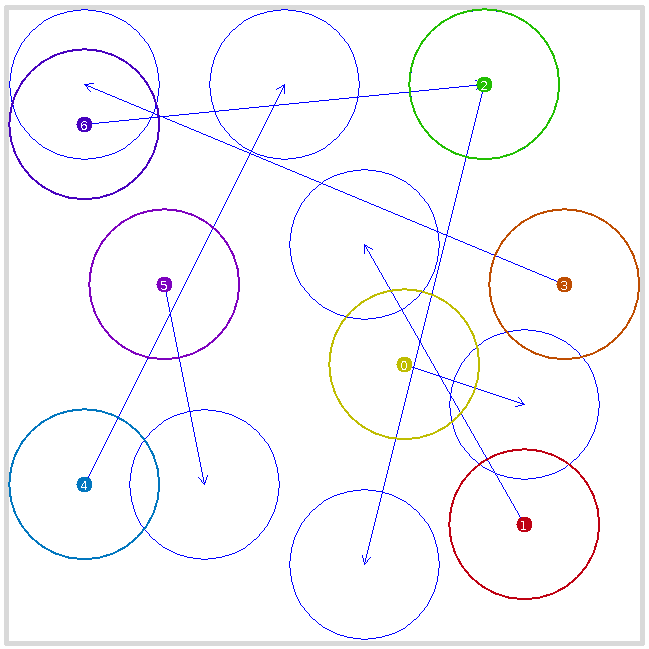} & \includegraphics[width=4cm]{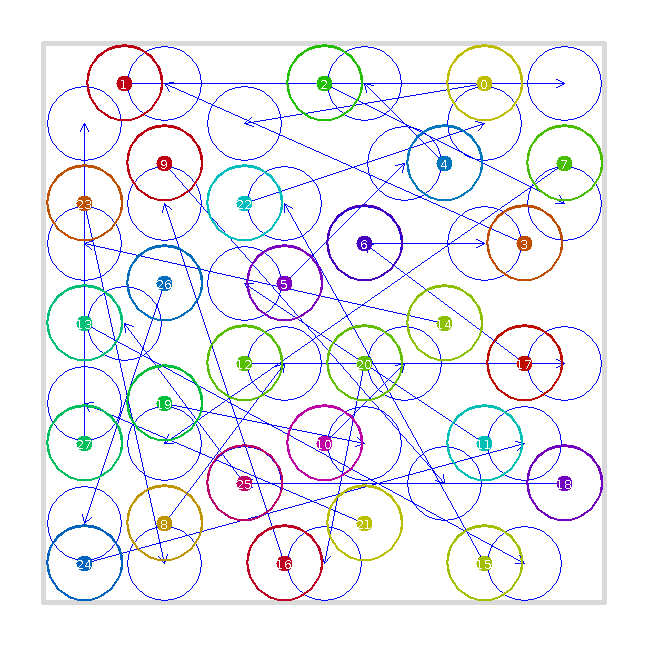} & \includegraphics[width=4cm]{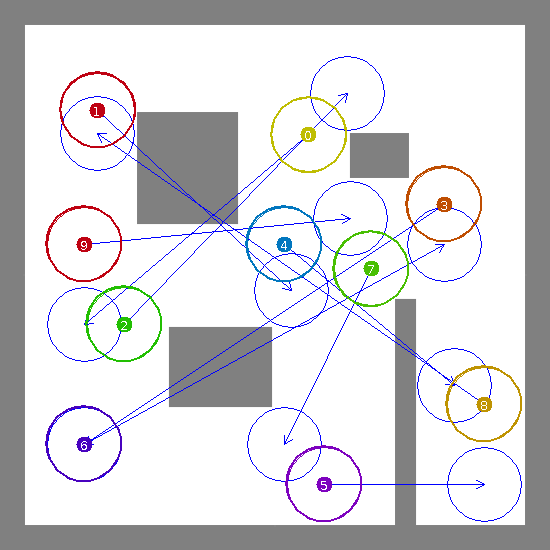}\tabularnewline
example instance, 7 robots  & example instance, 25 robots & example instance, 10 robots\tabularnewline
(a) Scenario A  & (b) Scenario B & (c) Scenario C\tabularnewline
\end{tabular} 
\par\end{centering}

\caption{\label{fig:Environments}Experimental environments}
\end{figure*}

\subsection*{Experiment setup}

\begin{figure}
\begin{centering}
\subfloat[Discretization used in Scenario A]{\begin{centering}
\includegraphics[width=4cm]{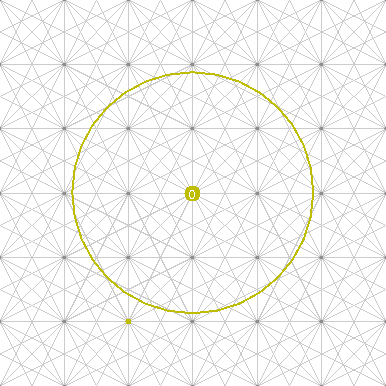}
\par\end{centering}

}\subfloat[Discretization used in Scenario B and C]{\begin{centering}
\includegraphics[width=4cm]{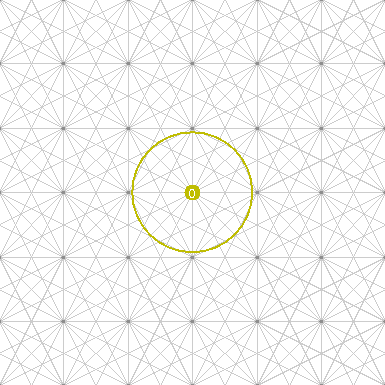}
\par\end{centering}

}
\par\end{centering}

\caption{\label{fig:Discretization}Discretization used for trajectory planning}
\end{figure}

The experimental comparison is done in three environments\emph{ }depicted
in Figure~\ref{fig:Environments}. The robots are modeled as 2-d
discs that move on a 16-connected grid graph depicted in Figure~\ref{fig:Discretization}.
The start and goal position for each robot is chosen randomly. When
generating start and destination for each robot we ensure that a)
robots do not overlap at start position, b) robots do not overlap
at goal position, and c) when adding a robot, its individually optimal
trajectory must be in conflict with the trajectory of some previously
added robot, i.e. the robots form a single conflict cluster. In Scenario
A and B, the start and destination for each robot is chosen from the
area depicted by the gray rectangle, however, the robots are allowed
to leave the rectangle when resolving the conflict. In \emph{Scenario
C}, the robots cannot leave the free space when resolving the conflict.
All robots can move at the same maximum speed.

We generated 25 random problem instances for different numbers of
robots in each environment. On each instance we run all the algorithms
and record the solution quality of the returned solution and the runtime.
The bars in the graphs indicate standard error of the average. Experiment
runs were executed on 1 core of Intel Xeon E5-2665 2.40GHz, 4 GB RAM.
A bundle that includes implementation of all the algorithms together
with the problem instances can be downloaded at http://agents.fel.cvut.cz/\textasciitilde{}cap/kpm.

\subsection*{Algorithms used in the comparison}

\paragraph*{Penalty Method (PM)}

The penalty-based method described in Section~\ref{sec:Algorithm}
was tried for different values of parameters $k$ ranging from $k=3$
to $k=100$.

\paragraph*{Prioritized Planning (PP)}

In prioritized planning (PP) we use a fixed random priority ordering
over the robots. The ordering used in PP identical to the ordering
used in the penalty method.

\paragraph*{Operator Decomposition (OD)}

Operator Decomposition~\cite{Standley10} is a complete and optimal
forward-search algorithm for multi-robot path planning on a graph
representing a discretization of the joint state space of a number
of robots. The joint state space is searched using the operator decomposition
technique that decomposes joint-actions to trees of single-robot moves,
which allows more efficient pruning and better utilization of a heuristic
estimate during the search process. We use OD algorithm to find provably
optimal solutions.

\paragraph*{Optimal Reciprocal Collision Avoidance (ORCA)}

The reactive technique ORCA~\cite{vanBerg2011_ORCA} is typically
used as a closed-loop controller that at each time instant selects
collision-avoiding velocity vector from the continuous space of robot's
velocities that is the closest to the robot's desired velocity. In
our implementation, at each time instant the algorithm computes a
shortest path from the robots current position to its goal on the
same graph that is used by other methods. The desired velocity vector
then points at the this shortest path at the maximum speed. When using
ORCA, we often witnessed dead-lock situations during which the robots
either moved at extremely slow velocity or stopped completely. If
a prolonged deadlock situation was detected, we considered the run
as failed.

\subsection*{Results}

\paragraph*{Success rate}

\begin{figure}
\begin{centering}
\includegraphics[scale=0.7]{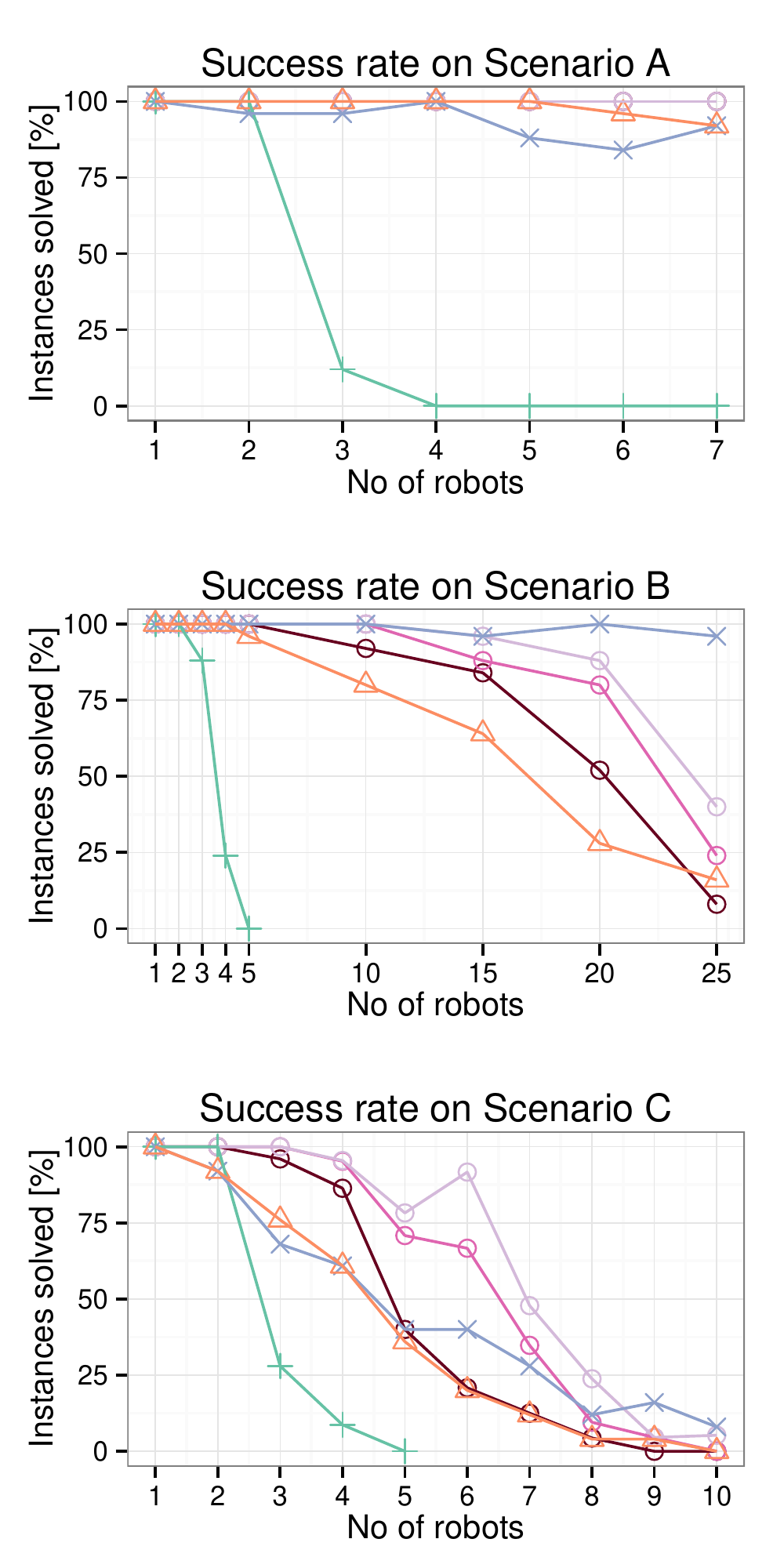}
\par\end{centering}

\begin{centering}
\includegraphics[scale=0.7]{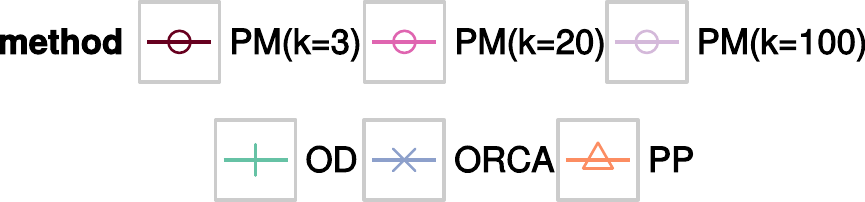}
\par\end{centering}

\caption{\label{fig:Success}Success rate. The plot shows the percentage of
instances successfully solved by PM, PP, ORCA and OD in each environment.
From PM, three different values of $k$ parameter are considered:
$k=3$, $k=20$, and $k=100$. }
\end{figure}

The comparison of success rate of the tested algorithms is in Figure~
\ref{fig:Success}. The plots shows the ratio of successfully solved
instances for each algorithm in each test environment. Since the runtime
of OD grows exponentially, we limited the runtime of OD to 1 hour.
The simulation of ORCA was terminated with failure if a deadlock was
detected.

\paragraph*{Sub-optimality}

\begin{figure}
\begin{centering}
\includegraphics[width=0.9\columnwidth]{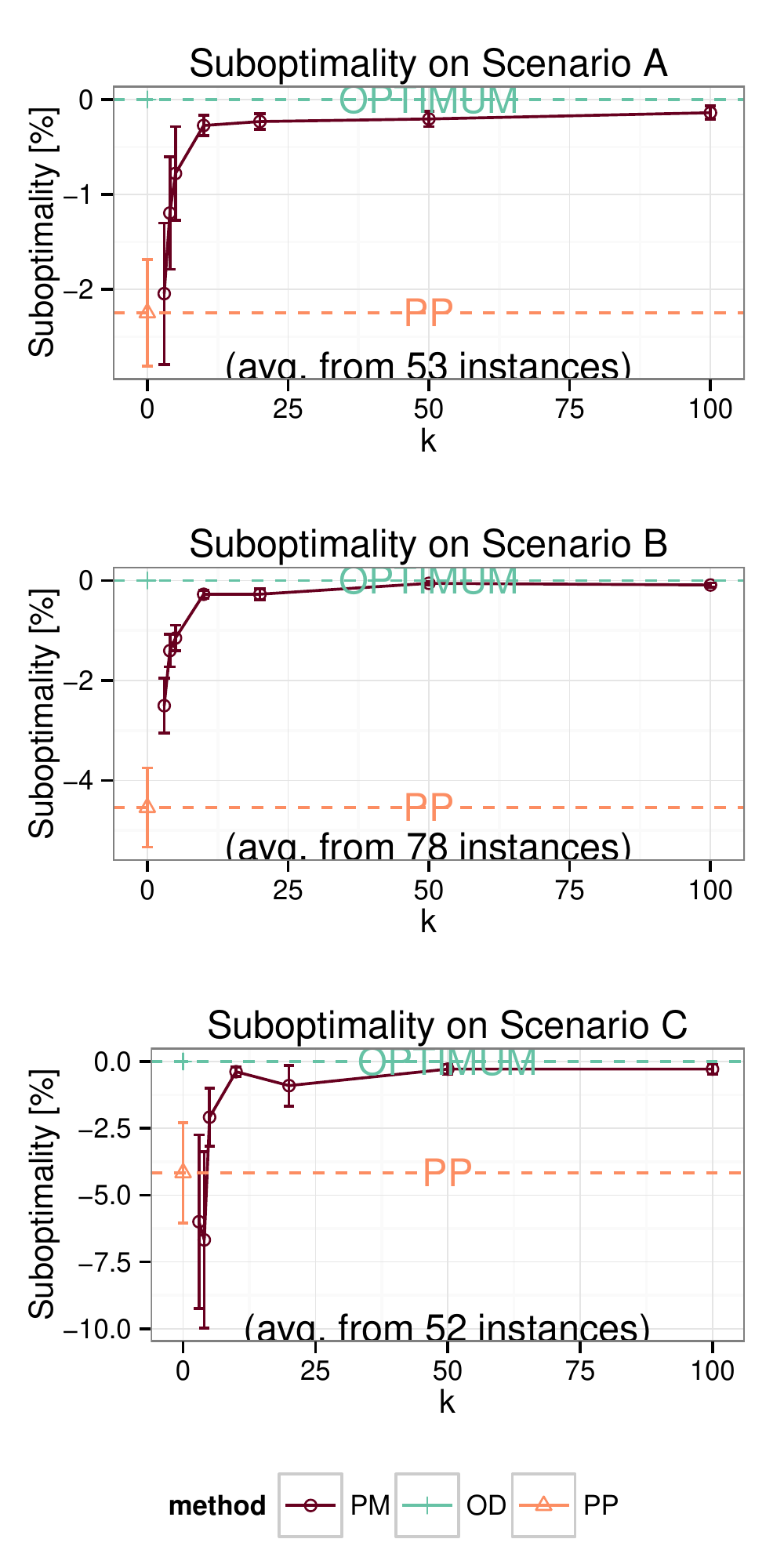}
\par\end{centering}

\caption{\label{fig:Suboptimality}Optimality of penalty method. The plot shows
average suboptimality of solutions returned by PM for different values
of $k$ parameter ranging from $k=3$ to $k=100$. The averages are
computed on instances for which we were able to compute the optimum
using the OD algorithm.}
\end{figure}

Figure~\ref{fig:Suboptimality} shows average sub-optimality of solutions
returned by the PM algorithm for different values of $k$ parameter
in each environment. The average suboptimality of solutions returned
by PP algorithm is also shown for each environment. The sub-optimality
of a solution on a particular instance is computed as $-(c-c^{*})/c^{*}$,
where $c$ is the cost of the solution returned by the measured algorithm
and $c^{*}$is the cost of optimal solution computed using OD algorithm.
The averages are computed only from the subset of instances that were
successfully resolved by all tested algorithms and for which we were
able to compute the optimum.

\paragraph*{Time spent outside goal}

\begin{figure}[h]
\begin{centering}
\includegraphics[width=0.9\columnwidth]{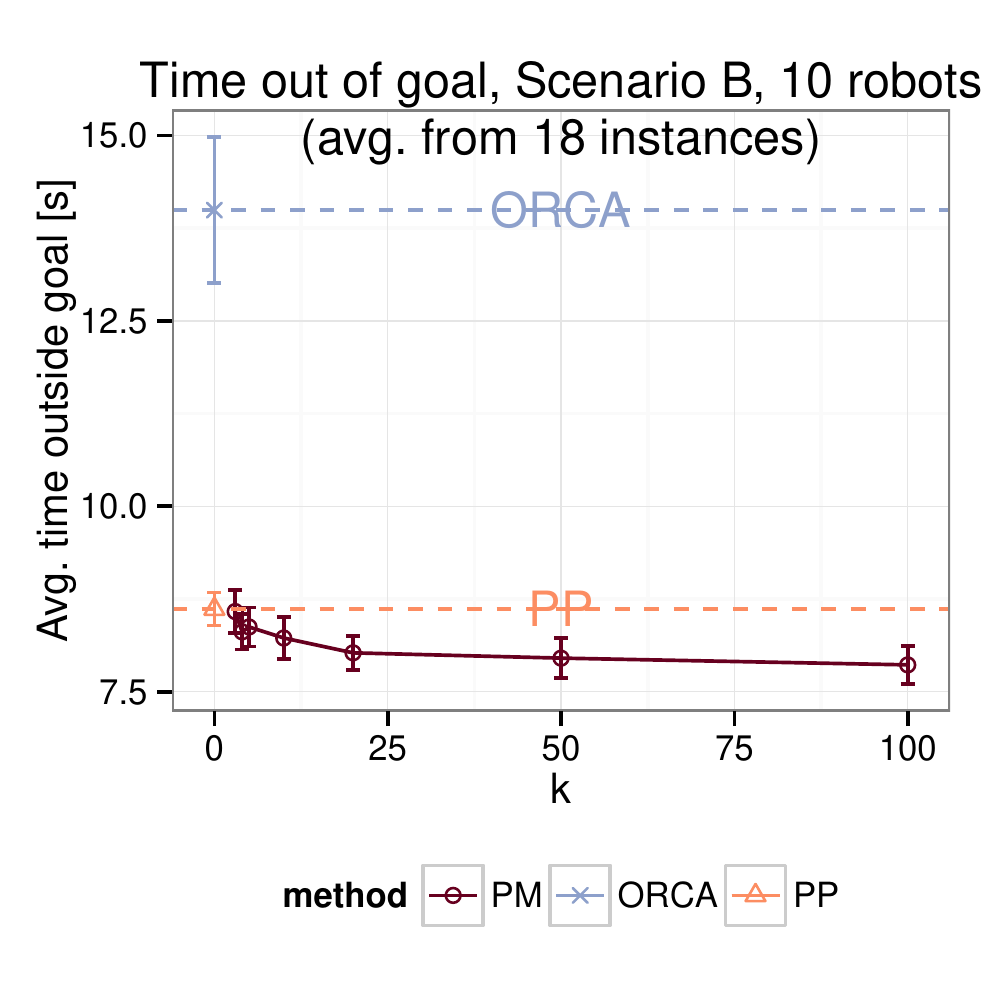}
\par\end{centering}

\caption{\label{fig:Quality}Average time spent outside goal (i.e. cost). The
plot shows average time the robots spend outside the goal if they
execute solutions found by PM, PP and ORCA. For PM, we consider values
of $k$ ranging from $k=3$ to$k=100$. Measured on all instances
with 10 robots in Scenario B that were successfully solved by all
compared methods.}
\end{figure}

Figure~\ref{fig:Quality} shows the average time spent outside goal
position when the robots execute the solutions found by PM, PP and
ORCA in all instances in Scenario B that involve 10 robots. The averages
are computed on the subset of instances that were successfully resolved
by all compared algorithms. 

\paragraph*{CPU runtime}

\begin{figure}[h]
\begin{centering}
\includegraphics[width=0.9\columnwidth]{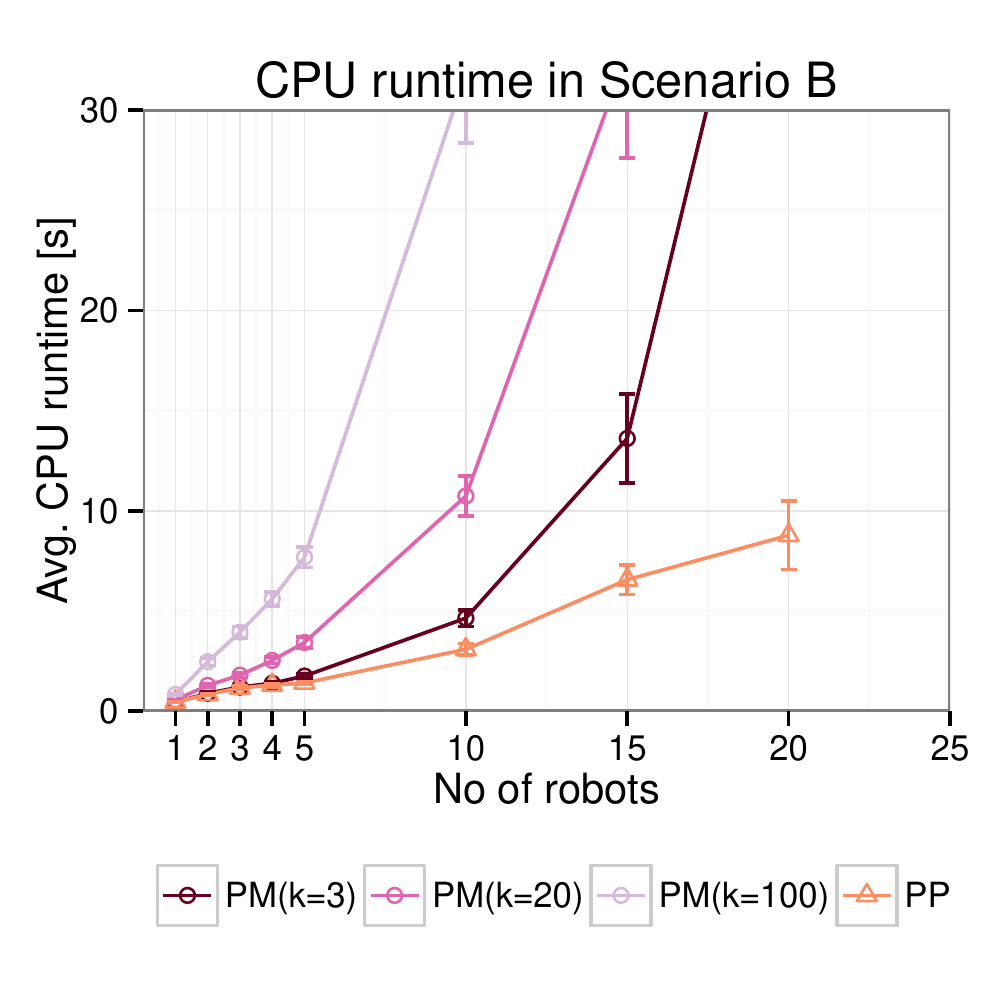}
\par\end{centering}

\caption{\label{fig:CPU-runtime}CPU runtime requirements. The plot shows average
CPU runtime needed by PM and PP to return solution in instances involving
different numbers of robots. From PM, three different values of $k$
parameter are considered: $k=3$, $k=20$, and $k=100$. }
\end{figure}

Figure~\ref{fig:CPU-runtime} shows average CPU runtime that PM and
PP algorithms require to return a solution. The averages are computed
only on the subset of instances that were successfully resolved by
both PP and PM. 

\subsection*{Results interpretation}

Our results show that the optimal algorithm OD is practical only for
instances that contain three or fewer robots. Prioritized planning
generally scales better than OD, but returns solutions that are on
average 2-4\,\% suboptimal (cf.~Figure~\ref{fig:Suboptimality}).
Although PM requires more time to provide a solution than PP (cf.~Figure~\ref{fig:CPU-runtime}),
it does not exhibit the exponential drop in ability to solve larger
instances that OD suffers from. Yet, the cost of solutions returned
by PM tend to converge to the vicinity of the optimal cost with increasing
number of iterations $k$ for the instances where the optimum is known,
(cf. Figure~\ref{fig:Suboptimality}) and otherwise converges to
solutions that are of significantly lower cost than the solutions
returned by PP (cf.~Figure~\ref{fig:Quality}). Further, besides
the improved quality of the returned solutions, PM achieves higher
success-rate on our instances (cf.~Figure~\ref{fig:Success}). E.g.,
in the empty environment with 20 robots (cf.~Figure~\ref{fig:Success}),
PM(k=100) solves 88\,\% of the instances, where PP solves only 28\,\%.

Although ORCA exhibits high success rate on our instances in Scenario
B, it should be noted that the trajectories resulting from this collision
avoidance process are very costly. We have observed that in instances
involving higher number of robots kPM returns trajectories that are
more than 40\,\% faster than the trajectories returned by ORCA (cf.~Figure~\ref{fig:Quality}).

\subsection*{Real-world maps}

\begin{figure}
\begin{centering}
\subfloat[\label{fig:Office-roadmap}Tasks of the robots and the roadmap used
for planning.]{\begin{centering}
\includegraphics[width=1\columnwidth]{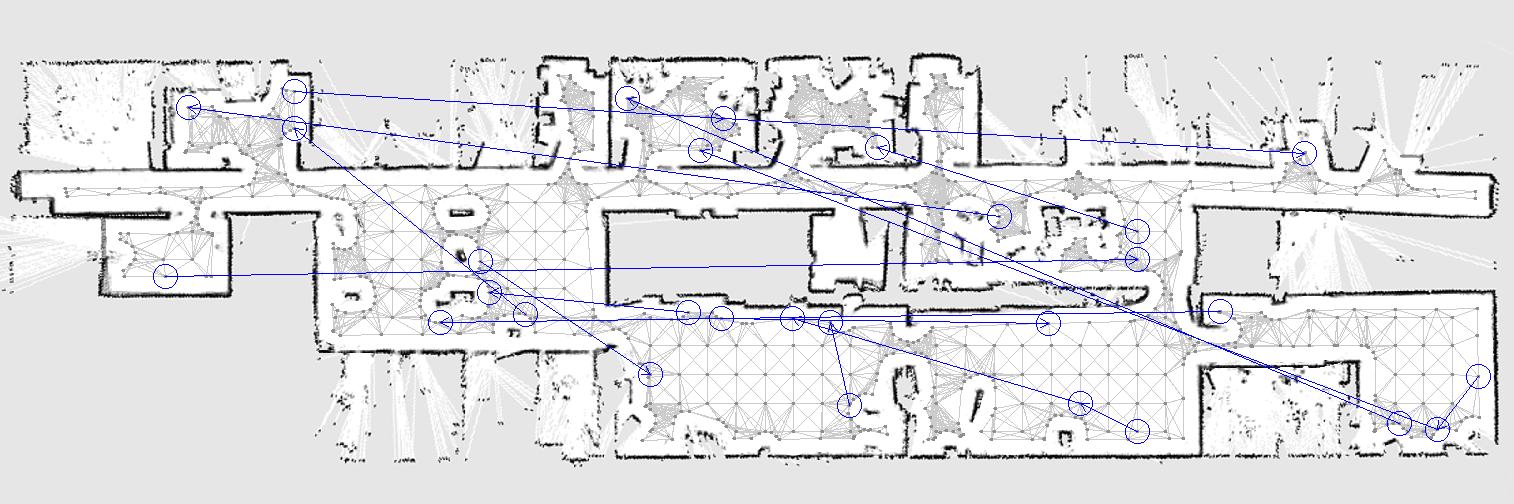}
\par\end{centering}

}~\\

\par\end{centering}

\begin{centering}
\subfloat[\label{fig:Office-results}Resulting coordinated trajectories.]{\begin{centering}
\includegraphics[width=1\columnwidth]{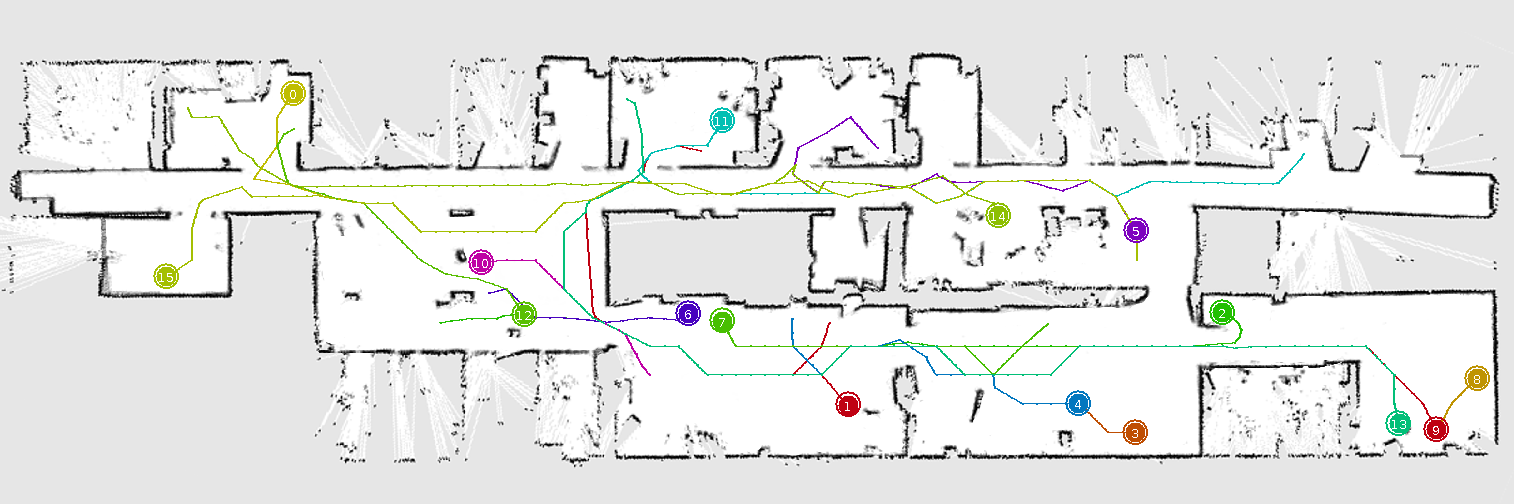}
\par\end{centering}

}
\par\end{centering}

\caption{\label{fig:Office-corridor}Office corridor: Coordination of 16 robots
in an office corridor.}
\end{figure}

To demonstrate the applicability of our method, we have deployed the
penalty method to coordinate trajectories of a number of simulated
robots in two representative real-world environments. First, we tested
the algorithm in an \emph{office corridor} environment, which is based
on the laser rangefinder log of Cartesium building at the University
of Bremen.%
\footnote{We thank Cyrill Stachniss for providing the data through the Robotics
Data Set Repository~\cite{Radish}.%
} The environment and the roadmap used for trajectory planning in the
environment together with the task of each robot are depicted in Figure~\ref{fig:Office-roadmap}.
We run PM(k=5) algorithm to coordinate the trajectories of 16 robots
sharing the environment and after 19 seconds obtained the trajectories
shown in Figure~\ref{fig:Office-results}. A video showing simulated
execution of the found trajectories can be watched at http://youtu.be/VfiBuQBBIhM. 

\begin{figure}
\begin{centering}
\subfloat[\label{fig:Warehouse-roadmap}Tasks of the robots and the roadmap
used for planning.]{\begin{centering}
\includegraphics[width=0.85\columnwidth]{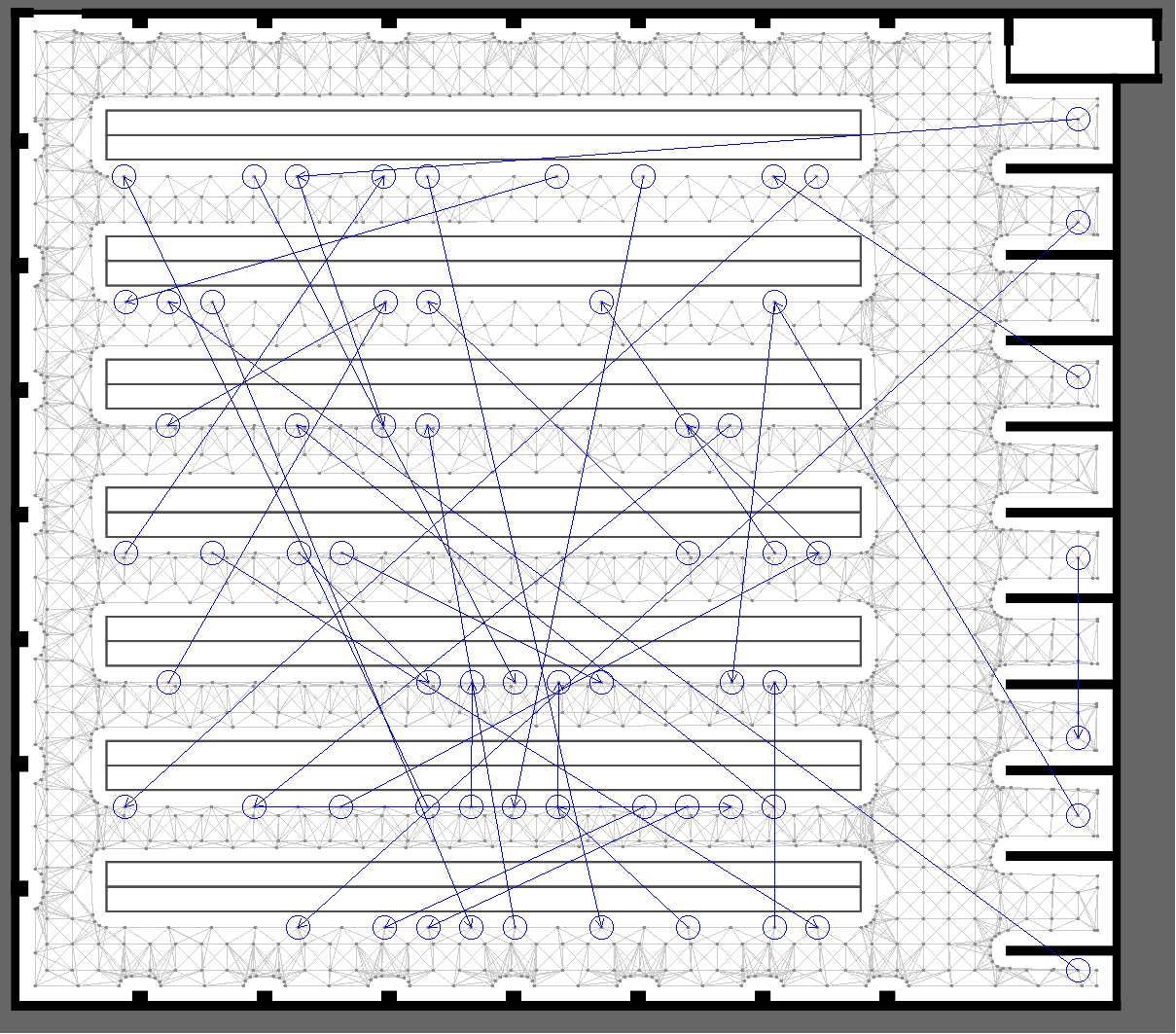}
\par\end{centering}

}\\
~\\

\par\end{centering}

\begin{centering}
\subfloat[\label{fig:Warehouse-result}Resulting coordinated trajectories.]{\begin{centering}
\includegraphics[width=0.85\columnwidth]{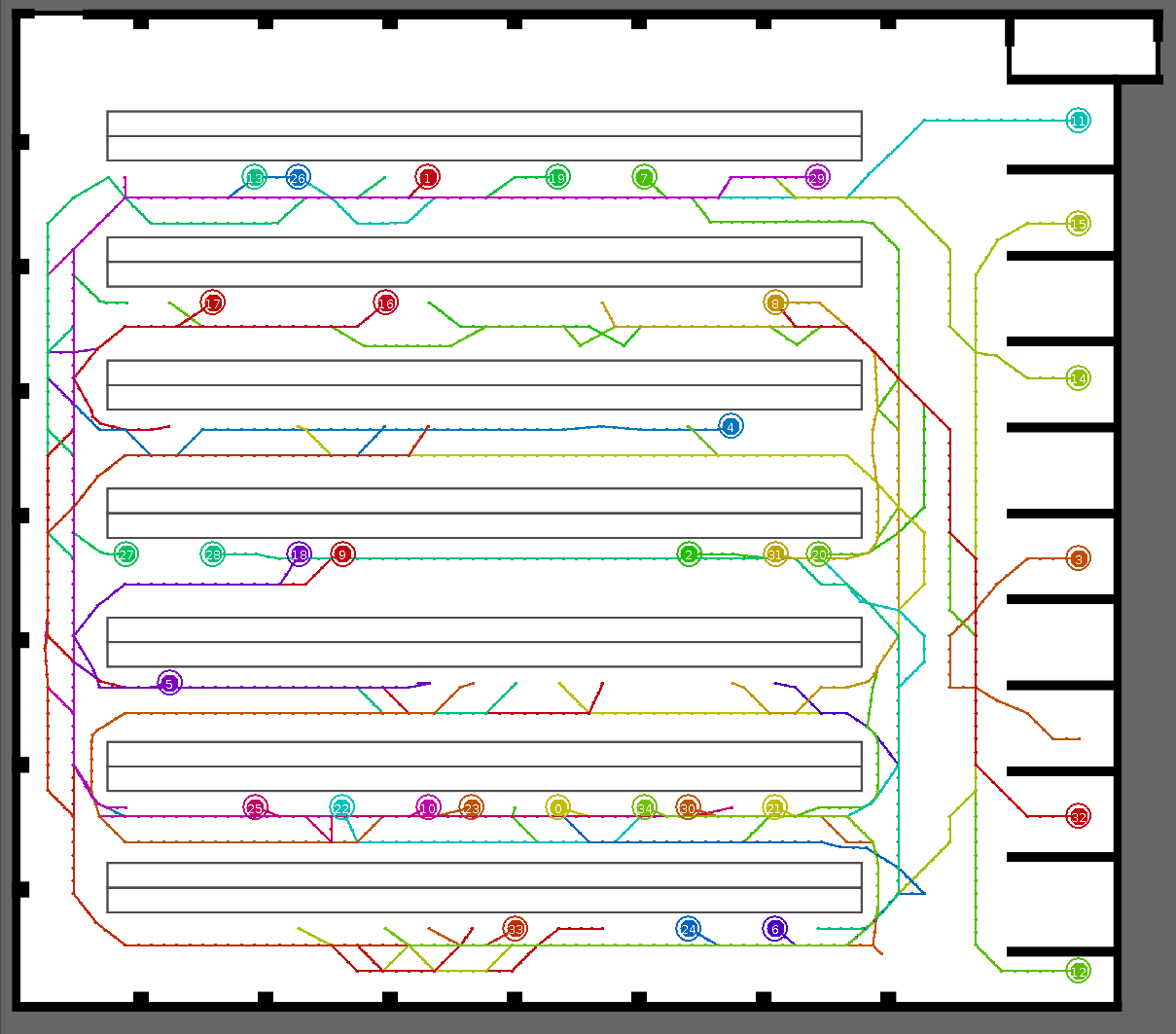}
\par\end{centering}

}
\par\end{centering}

\caption{\label{fig:Warehouse}Logistic center: Coordination of 35 robots in
a logistic centers.}
\end{figure}

Second, we depolyed the algorithm in a \emph{logistic center} environment.
The tasks of the individual robots and the roadmap that the robots
used for planning in this environment are shown in Figure~\ref{fig:Warehouse-roadmap}.
We used PM(k=5) algorithm to coordinate the trajectories of the individual
robots. After 55 seconds we obtained coordinated trajectories shown
in Figure~\ref{fig:Warehouse-result}. The simulated execution of
the found trajectories can be watched at http://youtu.be/G3A3TYKu73Q.

\section{\label{sec:Conclusion}Conclusion}

In this work we have explored the applicability of penalty-based method
to improve success rate and the quality of returned solution of prioritized
planning. We have formulated a new penalty-based algorithm for finding
collision-avoiding trajectories in multi-robot teams. The algorithm
starts from an individually optimal trajectory for each robot. Then,
the parts of the trajectories that lie in collision with other robots
are penalized and the magnitude of the penalty is gradually increased
towards infinity. After each such increase, the trajectory of one
of the robots is replanned to account for the increased penalty. Using
this process, the trajectories are gradually forced out of collisions. 

We have compared our penalty-based method with a state-of-the-art
optimal algorithm, prioritized planning and reactive technique ORCA
on three benchmark scenarios. Our results show that with increasing
number of iterations, the algorithm constructs solutions with near-optimal
cost (on instances where the optimum was known). On the instances
where the optimum was not known, our method consistently provided
solutions that are 4-10~\% cheaper than solutions provided by prioritized
planning and up to 40~\% cheaper than the solutions provided by a
widely-used reactive technique ORCA.

\bibliographystyle{plain}
\bibliography{bib}

\end{document}